# Analysis of communication patterns with scammers in Enron corpus

**Dinesh Balaji Sashikanth**

Master of Science in Computer Science

School of Informatics and Computing

Indiana University,Bloomington-47405,USA

## Abstract

This paper is an exploratory analysis into fraud detection taking Enron email corpus as the case study. The paper posits conclusions like strict servitude and unquestionable faith among employees as breeding grounds for sham among higher executives. We also try to infer on the nature of communication between fraudulent employees and between non-fraudulent-fraudulent employees

## 1 Rationale

This document provides an in-depth scrutiny of communication patterns between venal and genuine employees in the Enron email corpus. This project work is for the partial fulfillment for course B659 Detecting latent properties in text. I am thankful to Professor Markus Dickinson, Associate Professor in Linguistics, Indiana University for providing valuable suggestions and guidance.

## 2 Introduction

Enron was created in 1985 through the merger of two natural gas companies at the behest of Houston executive Kenneth Lay. Enron was an American energy conglomerate based in Houston, Texas. During the 1990s, it was considered one of the most powerful and successful corporations in the world. Lay remained the chief executive of Enron throughout its existence

Beginning in the late 1990s, Enron executives such as Jeffrey Skilling and Andrew Fastow initiated a campaign to hide business losses from company stockholders and the general public. Enron was subsequently granted – government deregulation. As a result of this declaration of deregulation, Enron executives were permitted to maintain agency over the earnings reports that were released to investors and employees alike.

2.1 Deceitful Practices

Enron and other energy suppliers earned profits by providing services such as wholesale trading and risk management in addition to building and maintaining electric power plants, natural gas pipelines, storage, and processing facilities. Instead of reporting the profit on the trading and brokerage fees, Enron chose to report the entire value of each of its trades as revenue.

In the beginning, the company listed the cost of supplying energy and the revenue it obtained from that. But after Skilling joined the company, he adopted the mark to market accounting practice. Mark-to-market accounting requires that once a long-term contract was signed, income is estimated as the present value of net future cash flow.

Enron used special purpose entities—limited partnerships or companies created to fulfill a temporary or specific purpose to fund or manage risks associated with specific assets.

The special purpose entities were used for more than just circumventing accounting conventions. Enron's balance sheet understated its liabilities and overstated its equity, and its earnings were overstated. Enron disclosed to its shareholders that it had hedged downside risk in its own illiquid investments using special purpose entities. However, investors were oblivious to the fact that the special purpose entities were actually using the company's own stock and financial guarantees to finance these hedges. This prevented Enron from being protected from the downside risk. Notable examples of special purpose entities that Enron employed were JEDI, Chewco, Whitewing, and LJM.

2.2 Consequences

The scandal, revealed in October 2001, eventually led to the bankruptcy of the Enron Corporation. the company's stock price, which achieved a high of US$90.75 per share in mid-2000, plummeted to less than $1 by the end of November 2001. Enron's shareholders lost $74 billion in the four years before the company's bankruptcy ($40 to $45 billion was attributed to fraud).As Enron had nearly $67 billion that it owed creditors, employees and shareholders received limited, if any, assistance aside from severance from Enron. Many executives at Enron were indicted for a variety of charges and some were later sentenced to prison.

2.3 Enron email corpus

The Enron Corpus is a large database of over 600,000 emails generated by 158 employees of the Enron Corporation and acquired by the Federal Energy Regulatory Commission during its investigation after the company's collapse. A copy of the database was subsequently purchased and released by Andrew McCallum, a computer scientist at the University of Massachusetts Amherst. The corpus is "unique" in that it is one of the only publicly available mass collections of "real" emails easily available for study , as such collections are typically bound by numerous privacy and legal restrictions which render them prohibitively difficult to access.

## 3 Previous work

Some of the well-known research works on this corpus were: Research on dynamics of the structure and properties of organizational communication network as a function of time (Diesner et al 2005), detecting community structure based on link mining (Qian et al 2006).
From the linguistics perspective-work on formal communication as a factor of social distance, relative power and weight of imposition (Peterson et al 2011), Recall enhancing methods for named entity recognition with Enron email as case study (Minkov et al 2005) and exploring the use of word 'virtual' in Enron corpus to establish the polysemic usage in contemporary contexts (Greg et al 2009)

## 4 Research Motivations

The primary motivation for this research was to study and detect patterns in fraudulent communication by using existing research in deception detection and author profiling. The Enron email corpus was adopted because it represented a true large-scale-real world communication where the emails of employees were available for research to the public without any privacy concern and also the ease of labelling fraudulent employees using the available legal records data and extensively studying their communication patterns. But this dataset proved to be an ill-fitting for the research (extensively addressed in section 5) because of which the primary goals of the research were readopted and are stated as follows **(S1 Genuine employees and S2 Fraudulent employees):**

*-(TASK 1/3) Analyze the communication patterns of S1 with S2( closed or open, transpar-*

ent or opaque, formal or casual so as to infer the level of trust S1 had over S2 and the level of transparency as perceived by S1 over S2)

-(TASK 2/3)How did S2 carry out their malice practice.(To learn about their modus operandi especially how open were they in communicating these plans.)

-(TASK 3/3)Communication network analysis of S2 across 2 different time intervals (Learn about significant changes in the communication structure when covert practices were at their peak and when these practices were being exposed

## 5 Datasets-The problems and splits

The dataset was acquired from CALO project hosted by Artificial Intelligence Center of SRI international and maintained by Carnegie Mellon University, Pennsylvania. ([www.cs.cmu.edu/enron](www.cs.cmu.edu/enron)). The dataset comprised of 619,446 messages of 158 users. Compiling from the internet, list of fraudulent employees revealed that only 4 out of the 158 employees were actually fraudulent. Data about the rest of the fraudulent employees were not made public. Hence our research focuses primarily on these 4 employees: Kenneth Lay, Chairman; Jeffery Skilling CEO; John Forney, Executive for energy trade and David Delainey, Head for trade and retail energy units.

As the dataset was a real-time collection, there was also the problem of spams, forwarded messages, discussion threads and mailing lists. While spams were outright not useful for this type of analysis task, forwarded messages and discussion threads were rejected because they were basically a series of messages where one party's communication pattern was influenced by another. Similarly group mails were not intended for a particular individual and hence were rejected.

Some peculiar problems revealed well into research and these were as follows: Kenneth Lay's sent messages were actually that from his assistant Rosalee fleming, Jeff Skilling's sent messages were that from his assistant Sherri sera. Over 30% of Jeff's inbox messages were about a donation drive and how much each employee pledged to contribute. David Delainey had only 50 messages in his inbox and the rest 1350 messages were in the form of discussion threads.

Due to the aforementioned problems, data for most of the tasks was collected manually(going over through the emails of these 4 employees and removing emails not meeting the criteria).Also only the received emails for these four were considered as their sent messages were not theirs or their primary response was through discussion thread. Though extreme care was taken in terms of data purity, the unavailability of large corpus particularly suited for this task renders the research inferences questionable for extrapolation to general instances.

For Task-1: 290 emails (103 Lay,43 JFor,93 JSki, 51 DDel) were considered

For Task-2: All of Inbox messages of the 4 employees were considered

For Task-3: emails from Jeff (inbox, notes_inbox, deleted items), Lay (inbox, note_inbox, deleted items), Forney (inbox, deleted items), Delainey (all_documents, inbox, notes_inbox, deleted items) were considered

## 6 Task-1/3 S1-S2 Communication

### 6.1 Methodology

The emails specifically for this task were hand-chosen. Although the expected analysis should be on the 2 way communication, because of the constraints mentioned above, only S1's communication with S2 was considered.

This implies that only the emails in inbox folders of fraudulent employees were taken into consideration. The emails considered were free of newsletters, discussion threads and charity drive messages. 290 emails were obtained which were used for analysis.

### 6.2 Analysis

The following parameters were evaluated:

**Formality**:-Francis et al 1999 describe that non-deictic category of words are used in formal contexts and these include noun, adjectives, prepositions and articles. The deictic category whose frequency decreases with increase in formality include pronouns, verbs, adverbs and interjection.

**F**= *(noun frequency +adjective frequency +preposition frequency +article frequency-pronoun frequency-verb frequency-adverb frequency-interjection frequency+100)/2*

The pos frequency is the number of times the pos tags appear over the total number of words.

| Categories | F score | Categories | F score |
|---|---|---|---|
| Phone conversation | 36 | Prepared speeches | 50 |
| Spontaneous speeches | 44 | Broadcasts | 55 |
| interviews | 46 | Writing | 58 |
| Imaginative writing | 47 | Informational writing | 61 |

Table-1 F scores for different outlets in English (Francis et al)

**Diversity**: also defined as the type token ratio (miller et al 1981)

*Diversity= (Total number of different words or terms/Total number of words and terms)*

Original proposed as a measure of child language development (Brian,1986), research( Victoria et al) has shown that written text has more lexical diversity than oral text and if the lexical diversity is less than 50% for any form of communication, it implies ambiguity or unnatural communication pattern.

**Emotiveness**:

Proposed by Zhou et al 2004, emotiveness is defined as the expressivity of a language.

*Emotiveness=(Number of adjectives+Number of adverbs)/(Number of nouns+Number of verbs)*

Though less documented, Zhou et al show that the values to typically range from 0.25-0.35 for written conversation texts.

**LIWC measures**

The free online version of LIWC(Linguistic inquiry word count) was used to measure the following dimensions: *self-references, social words, positive emotions, negative emotions, overall cognitive words, articles and big words*. The LIWC text processing module reads each target word and compares with its dictionaries and if a match is found, the appropriate word category is incremented. Thus LIWC dimension capture basic emotional and cognitive dimension of personality. One interesting feature that makes it applicable in our research is its comparison of the scores in both formal and informal contexts

### 6.3 Results

Table-2 Formality score in communicating With Fraudulent employees

| Employee | Formality score(in %) |
|---|---|
| Kenneth Lay | 69.32 |
| Jeff Skilling | 70.31 |
| David Delainey | 73.23 |
| John Forney | 66 |

Table-3 Emotiveness measures in communicating With Fraudulent employees

| Employee | Emotiveness Score |
|---|---|
| Kenneth Lay | 0.22 |
| Jeff Skilling | 0.236 |
| David Delainey | 0.193 |
| John Forney | 0.216 |

Table-4 Diversity measures in communicating With Fraudulent employees

| Employee | Diversity Score |
|---|---|
| Kenneth Lay | 0.276 |
| Jeff Skilling | 0.293 |
| David Delainey | 0.300 |
| John Forney | 0.337 |

Table-5 LIWC dimensions in communicating With Fraudulent employees

|  | Formal contexts | Informal contexts | K.Lay | J.Ski | D.Del | J.For |
|---|---|---|---|---|---|---|
| Self-references | 4.2 | 11.4 | 4.07 | 4.27 | 5.05 | 4.33 |
| Social words | 8.0 | 9.5 | 8.70 | 9.86 | 7.32 | 7.60 |
| Positive emotions | 2.6 | 2.7 | 3.04 | 3.35 | 2.72 | 2.18 |
| Negative emotions | 1.6 | 2.6 | 0.31 | 0.46 | 0.38 | 0.16 |
| Overall cognitive words | 5.4 | 7.8 | 4.68 | 5.37 | 6.36 | 6.60 |
| articles | 7.2 | 5.0 | 6.36 | 6.15 | 5.83 | 6.16 |
| Big words | 19.6 | 13.1 | 25.14 | 24.72 | 21.62 | 18.46 |

The high formality and low emotiveness and diversity scores indicate that the correspondence other employees had with these fraudulent employees was highly solemn. Given these men were high rankings in Enron, these scores could indicate that other employees had no knowledge about the scam and had to comply by their decisions. Although we don't expect these figures to vary in case of genuine higher ups, this does indicate a conducive environment for formation of clique of fraudulent employees given the unquestionable faith other employees had on them. The result definitely rules out any possibility of conflict of interest in shaping up of the scam had other employees had any knowledge about it.

## 7 Task 2/3 Intra S2 communications

### 7.1 Methodology

To accurately learn about the conversations that the fraudulent employees had with other employees, it is often a good strategy to assume the conversation of a particular type and look at corpus to see what percentage of the corpus reinforces our assumption. Hence the following strategy was followed:

-A flaglist of words for possible indication of scam was created. This included words like*{bribe,bribed,corrupt,corrupted,scam,kick back,graft,gift,legal,audit,incentive,jedi,chewco,hedge funds…}*.The flaglist comprised of 40 words in total.

-The emails in the inboxes of these four people which qualify the non-forwarded, non-junk criteria were considered.

-The body of the message is taken and each word in the message is compared with the words in the flaglist.

-If there is a word match then the email name is displayed else the next email is considered.

## 7.2 Results

A number of emails in inboxes were flagged because they consisted of words 'gift' or 'audit' or 'incentive' or 'legal' but these turned out to be false positives and hence excluding these 4 words, the emails consisting of other words were analyzed.

The results were dismal, only 1 email was found which indicated conspiracy: An email from Jeffery Sherrick (not convicted?) to David Delainey on manipulation of balance sheet. Some other less relevant emails were flagged like 2 employees bribing senior staff in Peurto Rico, on UC/CSU (California State University) suing Enron for breach of contract and about some employees threatening a female employee. Also newspaper articles revealing the financial frauds were widely circulated internally.

The only inference possible from this analysis was that the scammers never discussed their action plan over emails. As they were high valued and envied people, they would not have communicated their plans over company's email id for the fear of being tracked by federal government or their rivals.

## 8 Task 3/3 Communication network analysis as a function of time

8.1 Methodology

As Described in the dataset section, I have considered emails from their inboxes and also from their deleted items. The same criteria of non-junk, non-forwarded message were applied here. For analysis of network as a function of time, I made a binary split of '**Incline**' period and '**Decline**' period. The date July 1st 2001 was chosen as the cutoff because Enron rose to the summit and was crowned as 'the most innovative company' till 2001. Although Jan 2001 to June 2001 was crappy period for Enron where some senior executives resigned and stock prices dipped by few percentage, It was during July-September 2001 that Enron end began, during August 2001, Jeff resigned from Enron, news outlets began to expose the scam and the stock prices nose-dived. Hence the split was a binary split and aptly named 'Incline' of Enron and 'Decline' of Enron. For analysis, the X-From address and X-To address were removed and based on the date of the email, the addresses were placed into their respective bins. This data was fed as input to **Gephi** which is a network visualization and analysis tool.

8.2 Results

Fig-1 Communication Network Visualization for Incline period

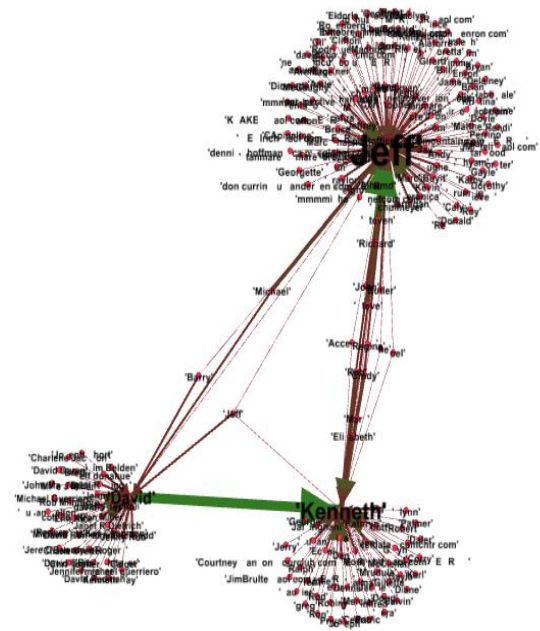

The above figure provides insights into the communication patterns other employees had with these 4 employees. There are 3 communities with 1 hub each: Kenneth, Jeff and David. Surprisingly Forney occurs as a peripheral node[between Kenneth and Jeff]. Jeff (CEO) played a major role interacting with a lot more employees than Kenneth Lay (Chairman). We also see the effect of allegiance of a set of employees to a particular executive.

The 'decline' graph presents a contrasting picture. Here Kenneth Lay is more dominant than Jeff Skilling. This structure reaffirms the fact of Jeff Skilling's resignation during this period and Kenneth Lay taking over as chairman and CEO of Enron. Surprisingly Forney[bottom left]is a hub and David delainey [not in picture] is a peripheral node.

Fig-1 Communication Network Visualization for decline period

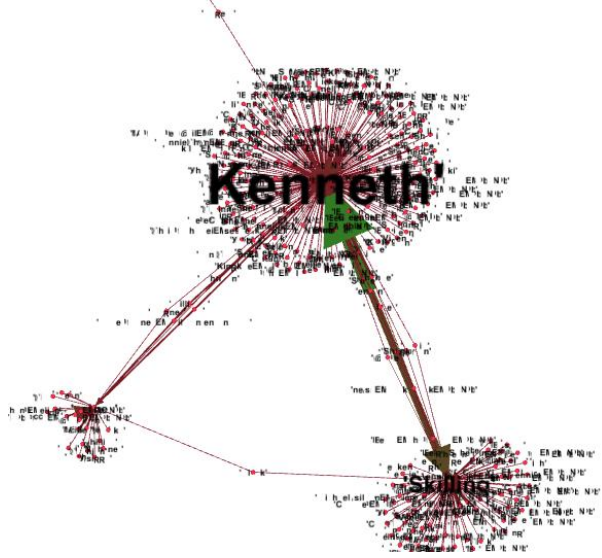

The high in-degree for Kenneth during 'decline' suggests that other employees looked up to him as the savior of their fates.

Table 6-Comparision of Network properties for the two graphs

|  | Incline | Decline |
| --- | --- | --- |
| Average clustering coefficient | 0.006 | 0 |
| Average weighted degree | 1.578 | 1.358 |
| Graph Density | 0.005 | 0.003 |

As I had considered only the inbox messages, the cluster coefficient is low because of the one way communication. Although the dominance in communication has changed significantly, the network properties are similar. The high affinity of employees to Kenneth Lay could explain for the decrease in weighted degree and the graph density.

## 9 Conclusions

Despite the dataset problems, this research has brought out accurately how other employees interacted with fraudulent employees. We now know that the genuine employees were serious in their communication, maintained and respected strict hierarchical order and never complained against the company's policies by higher executives. We also see that the fraudulent employees never communicated their plans over company's email for the fear of being exposed. While the conclusion cannot be generalized without a proper dataset and methodologies, maintaining a stronghold over subordinates, executing treason without a room for doubt and secretly colluding without any trace truly makes them *'The smartest guys in the room'(movie,2005)*

## 10 References

Wikipedia References

-Enron

-Enron scandal

www.nytimes.com/Enron timeline

www.efinancialnew.com/The Enron cast

www.USAtoday.com/A look at those involved...

*Automating linguistics based cues for detecting deception in text based asynchronous communication*, Lina Zhou et al, 2004,Group decision and negotiation 13,2004

*Using linguistic cues for automatic recognition of personalities in text* Mairesse et al, Journal of artificial intelligence,2007

*Formality of language: definition, measurement and behavioral determinants*, Francis et al , Internal report, University of Brussels,1999

*Type token ratio, what they really tell us?* Brian et al, Journal of child language,1987

*Analyzing language prediction in children: experimental procedures*- Miller 1981Baltimore University,Park press,